\documentclass[final,3p,times,twocolumn]{elsarticle}

\usepackage{amssymb}
\usepackage{amsmath}
\usepackage{comment}

\usepackage{multirow}
\usepackage{float}
\usepackage{booktabs}
\journal{Nuclear Physics B}

\begin{document}

\begin{frontmatter}



\author[1]{Kyle M. Regan \corref{cor1}}
\ead{regank@udel.edu}

\author[3]{Michael McLoughlin}
\ead{mike.mcloughlin@zeteotech.com}

\author[3]{Wayne A. Bryden}
\ead{wayne.bryden@zeteotech.com}

\author[2]{Gonzalo R. Arce}
\ead{arce@udel.edu}

\cortext[cor1]{Corresponding author}

\affiliation[1]{organization={Center for Bioinformatics and Computational Biology, University of Delaware},
             city={Newark},
             postcode={19713},
             state={Delaware},
             country={United States}}

\affiliation[2]{organization={Department of Electrical and Computer Engineering, University of Delaware},
             city={Newark},
             postcode={19716},
             state={Delaware},
             country={United States}}
\affiliation[3]{organization={Zeteo Tech Inc.},
             city={Sykesville},
             postcode={21784},
             state={Maryland},
             country={United States}}

\title{Unmasking Airborne Threats: Guided-Transformers for Portable Aerosol Mass Spectrometry\tnoteref{preprintnote}}
\tnotetext[preprintnote]{This preprint has been submitted to \textit{Computer in Biology and Medicine}.}




\begin{abstract}
Matrix Assisted Laser Desorption/Ionization Mass Spectrometry (MALDI-MS) is essential for biomolecular analysis in human health protection, offering precise identification of pathogens through unique mass spectral signatures to support environmental monitoring and disease prevention. However, traditional MALDI-MS relies on labor-intensive sample preparation and multi-shot spectral averaging, confining it to laboratory settings and hindering its application in dynamic, real-world settings critical for public health surveillance. These limitations are amplified in emerging portable aerosol MALDI-MS systems, where autonomous sampling produces noisy, single-shot spectra from a mixture of aerosol analytes, demanding novel computational detection methods to safeguard against infectious diseases. To address this, we introduce the Mass Spectral Dictionary-Guided Transformer (MS-DGFormer): a computational framework that processes raw, minimally prepared mass spectral data for accurate multi-label classification, directly enhancing real-time pathogen monitoring. Utilizing a transformer architecture to model long-range dependencies in spectral time-series, MS-DGFormer incorporates a novel dictionary encoder with Singular Value Decomposition (SVD)-derived denoised side information, empowering the model to extract vital biomolecular patterns from noisy single-shot spectra with high reliability. This approach achieves robust spectral identification in aerosol samples, supporting autonomous, real-time analysis in field-deployable systems. By reducing preprocessing requirements, MS-DGFormer facilitates portable MALDI-MS deployment in high-risk areas like public spaces, transforming human health strategies through proactive environmental monitoring, early detection of biological threats, and rapid response to mitigate disease spread.\\

\vspace{0.5em}
\noindent
\textbf{Preprint Notice.} This manuscript is a preprint and has been submitted to *Computer in Biology and Medicine*.
\end{abstract}


\end{frontmatter}



\section{Introduction}
\label{sec1}
Matrix Assisted Laser Desorption/Ionization Mass Spectrometry (MALDI-MS) has evolved to be a powerful tool to characterize and identify large biomolecules. MALDI-MS is a “soft ionization” method and typically utilizes a light absorbing 
matrix that, when mixed with an analytical sample and illuminated with a pulsed laser 
light,  will create ions from biomolecules to include proteins, peptides, and lipids \cite{ElAneed2009}. For biodefense applications, a major advantage of MALDI is that it can be combined with Time-of-Flight analyzers \cite{WileyMcLaren2004}, which do not require a complex fluidic system and can be miniaturized for field applications \cite{Bryden2000}. To achieve high mass resolution and accuracy, multiple methods such as delayed extraction and ion reflectors have been developed to compensate for energy spread during the ionization process \cite{Cotter2010}. Because MALDI-MS uses a sample deposited onto a substrate, deconvolution of a mixed sample can be quite complex and has motivated the development of single particle methods \cite{vanWuijckhuijse2005}, \cite{Papagiannopoulou2020}. A portable prototype MALDI-MS, introduced in ``digitalMALDI: A Single-Particle–Based Mass Spectrometric Detection System for Biomolecules'' \cite{Chen2025digitalMALDI} demonstrated the ability to produce spectra from environmental aerosol particles. The spectrometer autonomously samples aerosol particles, irradiates each by an ultraviolet laser creating ions from individual particles, then analyzes the ions by time-of-flight. The portability of this spectrometer enables in-field measurements for rapid identification of possible biological threats such as airborne bacteria, fungi, viruses, toxins, or nonvolatile chemicals. 

\begin{figure*}[t]
\centering
\includegraphics[width=\textwidth]{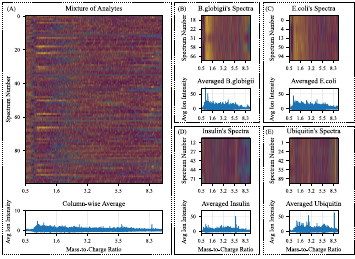}
\caption{(A) Top: An example batch of particles containing \(80\%\) dust particulate, with the remaining \(20\%\) evenly divided among the four biological markers. Each row represents a mass spectrum, and each column corresponds to a mass-to-charge ratio value. Bottom: The column-wise average. (B) Top: The heatmap of rows from (A) corresponding to \textit{B. globigii} spectra. Bottom: the average spectrum. (C), (D), (E) same format as (A) but with \textit{E. coli}, Insulin, and Ubiquitin, respectively.}\label{fig1}
\label{fig:fig1}
\end{figure*}

This research is driven by the urgent need to protect communities from natural and engineered biological threats, such as infectious outbreaks or bioterrorism. By enabling rapid detection of airborne pathogens, our work could curb disease spread and enhance public safety. Envision deploying compact devices in high-traffic areas like airports, transit systems, or stadiums for continuous, real-time scanning, to transform how we safeguard against environmental biological risks.

At the heart of this approach lies the near autonomous sampling of atmospheric *aerosols, a technique that enables continuous monitoring without relying on extensive human intervention. Yet, creating a threat detection system that is robust, accurate, and precise is not easy. False positives risk sparking unwarranted alarm, while false negatives could allow a silent spread of infection. Unlike controlled laboratory conditions with carefully prepared samples, environmental sampling captures a complex mixture of aerosol particles---primarily harmless background particles---making it challenging to identify the unique mass spectra of pathogens. This work tackles these challenges, laying the groundwork for a reliable system capable of safeguarding society from biological threats.

MALDI-MS is already FDA-cleared for identifying bacterial and fungal isolates in clinical labs, capable of distinguishing over 10,000 strains. While extensive research applies MALDI-MS to biological specimens, most focus on meticulously prepared lab samples requiring days of culturing or separation. Even then, single-laser-shot spectra are noisy, so multi-shot averaging is standard to boost signal-to-noise ratio before database matching.

This works well in clinics where shots target the same analyte, but environmental sampling involves diverse particles per shot. Averaging can blur features, as shown in Fig.~\ref{fig:fig1}, which depicts a batch of spectra from 80 dust aerosols mixed with five spectra each from \textit{Bacillus globigii}, \textit{E. coli}, Insulin, and Ubiquitin. Averaging assumes uniform analytes \cite{arce_mcloughlin}\cite{arce1986median}, but mixtures yield muddled spectra (Fig.~\ref{fig:fig1}A). A single-shot method is essential to isolate individual analyte spectra (Fig.~\ref{fig:fig1}B-E).

To enhance single-shot analysis, we leverage the low-rank structure of spectra using Singular Value Decomposition (SVD), a mathematical technique for denoising data by identifying dominant patterns. For a noiseless spectrum \(\mathbf{z} \in \mathbb{R}^{l}\) and noisy version \(\mathbf{s} = \mathbf{z} + \epsilon\), we form a matrix \(\mathbf{S} \in \mathbb{R}^{n \times l}\) from \(n\) spectra of the same class. SVD decomposes \(\mathbf{S} = \mathbf{U \Sigma V^T}\), where \(\mathbf{U}\in\mathbb{R}^{n \times n}\) and \(\mathbf{V}^{T}\in\mathbb{R}^{l \times l}\) are orthogonal matrices and \(\mathbf{\Sigma}\in\mathbb{R}^{n \times l}\) is a diagonal matrix containing the singular values in descending order. The matrix \(\mathbf{S}\) has a low-rank structure, with \(r \ll \min{(n,l)}\), indicating that \(\mathbf{z}\) resides within an \(r\)-dimensional subspace of \(\mathbb{R}^{l}\) spanned by the first \(r\) columns of \(\mathbf{V}\) \cite{Fessler_Nadakuditi_2024}. 
By retaining the top \(r\) singular values (with \(r \ll \min(n,l)\)) approximates the signal subspace, filtering out noise \cite{eckart1936approximation}.

This assumes uniform analytes, so for mixtures, we build a dictionary of denoised sub-dictionaries per analyte class via SVD, creating a union of subspaces as side information for feature extraction. We then employ a transformer encoder, a machine learning model adept at processing sequences like spectral peaks with positional context, to generate embeddings for input spectra and dictionaries separately. This separation allows spectral embeddings to extract biologically relevant features from the dictionary during training, and dictionary removal during inference, halving parameters for faster, deployable predictions crucial for field use in biological monitoring.





\section{Methods}
\label{sec2}
\subsection{Aerosol MALDI-MS Data Acquisition}
In this study, aerosol particles were ionized using ultraviolet (349 nm) laser pulses in a portable MALDI-Time-of-Flight (ToF) mass spectrometer, as detailed in our prototype system \cite{Chen2025digitalMALDI}. Each mass spectrum is represented as an intensity vector $\mathbf{s} = [s_1, s_2, \dots, s_l]^T \in \mathbb{R}^l$ paired with a corresponding $\frac{m}{z}$ vector $\mathbf{m} = [m_1, m_2, \dots, m_l]^T \in \mathbb{R}^l$. For batches of $n$ particles, the spectra form a matrix:

\[\mathbf{S} = \begin{bmatrix} s_{11} & s_{12} & \dots & s_{1l} \\ s_{21} & s_{22} & \dots & s_{2l} \\ \vdots & \vdots & \ddots & \vdots \\ s_{n1} & s_{n2} & \dots & s_{nl} \end{bmatrix} \in \mathbb{R}^{n \times l}\].

Since $\mathbf{m}$ is identical across measurements (unless machine parameters change), a two-dimensional $\mathbf{m}$ matrix is unnecessary. 

Due to the prototype phase, our data collection is limited to a handful of targets. To simulate field-expected spectral profiles for environmental pathogen monitoring, we selected bacterial (multi-peak, noisy), protein (few-peak), and non-biological (peakless) types. Data included two bacteria (\textit{Bacillus globigii}, \textit{Escherichia coli}), two proteins (insulin, ubiquitin) as positives, and Arizona Road Dust as negative background.
For safety, samples were aerosolized and collected in a lab setting, mimicking real-world conditions. Data acquisitions were performed in a class-specific manner, with \(\frac{m}z{}\) restricted to the range \([500, 10,000]\) Da. Each class produces a matrix of raw spectra, \(\mathbf{S}\), which is split into \(80\%\) for training and \(20\%\) for testing. Consequently, our model is trained on individual raw spectra, simulating single-shot detection, to ensure robust performance in field scenarios where a matrix of spectra may contain multiple, co-occurring classes. Table~\ref{tab:spectra_per_class} summarizes spectra counts and the training/testing split.

\begin{table}[h]
    \centering
    \caption{Number of Spectra per Class}
    \label{tab:spectra_per_class}
    \renewcommand{\arraystretch}{1.2} 
    \resizebox{\columnwidth}{!}{
    \begin{tabular}{|l|c|c|c|}
    
        \hline
        \textbf{Class} & \textbf{\# of Samples} & \textbf{Training Samples} &\textbf{Test Samples} \\
        \hline
        A.R.Dust& 630 & 504 & 126  \\
        \textit{B.globigii} & 1500 & 1200 & 300\\
        \textit{E.coli} & 1500 & 1200 & 300 \\
        Insulin & 1400& 1120 & 280 \\
        Ubiquitin & 1500 & 1200 & 300\\
        \hline
        \textbf{Total} & 6530 & 5224 & 1306 \\
        \hline
    \end{tabular}
    }
\end{table}
\subsection{Sparse Signal Processing}

A mass spectrum \(\mathbf{s}\) can be modeled by a linear combination of columns, referred to as atoms, from a dictionary matrix \(\mathbf{D} \in \mathbb{R}^{l \times \alpha}\), such that \(\mathbf{s}=\mathbf{D}\mathbf{x}\), where \(\mathbf{x} \in \mathbb{R}^{\alpha}\) is a sparse coefficient vector (\(\lvert\lvert \mathbf{x} \rvert\rvert_{0} \ll \alpha\)). This synthesis model leverages sparsity to denoise and extract relevant features from noisy spectra, crucial for real-time pathogen detection in complex aerosol signatures. 

Common dictionaries include orthogonal bases like Cosines, Fourier, or Wavelets (where \(l=\alpha\)), but over-complete dictionaries (\((l \ll \alpha)\)) \cite{mallat1993} allow more flexible representations, advancing applications in compressive sensing\cite{donoho}, dictionary learning\cite{Aharon2006}, medical imaging\cite{Ravishankar2011}, classification\cite{covid_CSEN}, and object detection\cite{Ahishali2023}. 

These approaches often solve the relaxed convex optimization problem known as Basis Pursuit (BP)\cite{chen2001atomic}:

\begin{equation}
            \min_{x}\; \lvert\lvert\mathbf{x}\rvert\rvert_{1} \text{ s.t. } \mathbf{s} = \mathbf{D}\mathbf{x}
            \label{eq:sparse_support}
    \end{equation}
\noindent

where the sparse vector, \(\mathbf{x}\), encodes rich task-specific information. This motivated data-driven dictionaries, as demonstrated in the Face Recognition imaging problem \cite{wright2009sparse}, where training samples form columns based on the principle that high-dimensional data from the same class lie in a low-dimensional subspace. A test sample is thus represented as a sparse linear combination of training samples via Sparse Representation Classification (SRC)\cite{wright2009sparse}. SRC has been applied to many applications such as image classification \cite{5206757}, denoising \cite{1014998}, and deep learning \cite{wen2016learningstructuredsparsitydeep}.

\subsection{Proposed Dictionary Construction}
\label{subsec:dictionary_construction}
Inspired by SRC, we constructed the dictionary $\mathbf{D} \in \mathbb{R}^{l \times \alpha}$ using $\alpha$ training spectra evenly distributed across $c$ classes ($\alpha/c$ per class), arranged column-wise with same-class spectra grouped into sub-dictionaries $\mathbf{D}_i$ ($i = 1, \dots, c$). Thus, $\mathbf{D}_i = [\mathbf{d}_{i,1}, \dots, \mathbf{d}_{i,\alpha/c}]$, and $\mathbf{D} = [\mathbf{D}_1, \dots, \mathbf{D}_c]$.

\begin{figure}[tb]
\centering
\includegraphics[width=0.95\columnwidth]{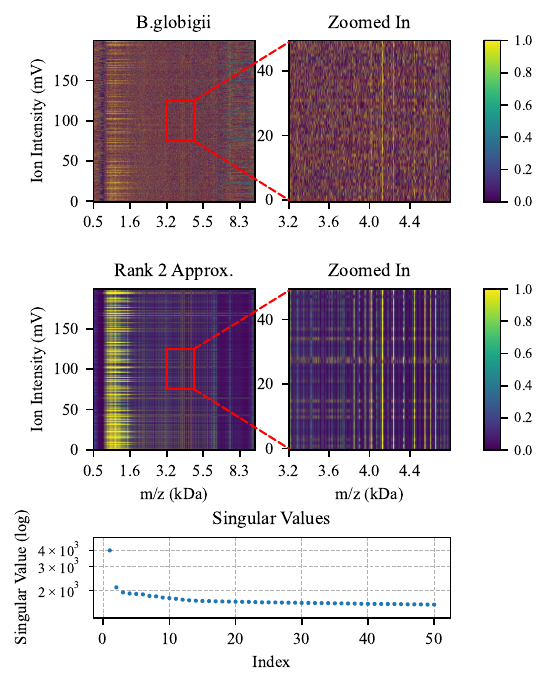}
\caption{Top: Heatmap of \(200\) mass spectra from \textit{Bacillus globigii}. Middle: A low-rank approximation via the Singular Value Decomposition (SVD) with rank \(r=2\). Bottom: The first 50 singular values from the SVD plotted on a y-axis log-scale.}
\label{fig:fig2}
\end{figure}

This class-specific clustering exploits the low-rank structure inherent in spectra from the same biomolecular class (rank $r \ll \alpha/c$). To denoise and capture essential features, we applied Singular Value Decomposition (SVD) to each sub-dictionary:
$$\tilde{\mathbf{D}}_i = \mathbf{U}_r \mathbf{\Sigma}_r \mathbf{V}_r^T, \quad r \ll \frac{\alpha}{c},$$
where $\mathbf{U}_r \in \mathbb{R}^{l \times r}$ and $\mathbf{V}_r \in \mathbb{R}^{\alpha/c \times r}$ are orthogonal matrices of left and right singular vectors, and $\mathbf{\Sigma}_r \in \mathbb{R}^{r \times r}$ contains the top $r$ singular values. The denoised dictionary was then formed by stacking these low-rank approximations:
$$\tilde{\mathbf{D}} = [\tilde{\mathbf{D}}_1, \tilde{\mathbf{D}}_2, \dots, \tilde{\mathbf{D}}_c] \in \mathbb{R}^{\alpha \times l}.$$
This creates a union of rank-$r$ subspaces that efficiently represent key biomolecular patterns in noisy spectra, enhancing multi-label classification for airborne pathogen detection in public health scenarios. The approach's efficacy is shown in Fig.~\ref{fig:fig2}, where SVD on \textit{Bacillus globigii} data reveals the first two singular values capturing primary features, with subsequent values as noise; a rank-2 approximation sharpens peaks and reduces noise for clearer identification.

\subsection{Transformers for Time-Series and Mass Spectrometry}
Transformers, first introduced for natural language processing \cite{vaswani2017attention}, have been extended to time-series analysis, including forecasting \cite{zhou2021informerefficienttransformerlong,zhou2022fedformerfrequencyenhanceddecomposed,garza2024timegpt1} and classification \cite{zerveas2020transformerbasedframeworkmultivariatetime,xu2022anomalytransformertimeseries}. In mass spectrometry (MS), they support peptide/protein identification and protein structure prediction \cite{Ekvall2022,yilmaz2024sequence}, capitalizing on long-range dependencies among spectral peaks. Typically, these models use denoised, high-resolution spectra from lab instruments, with $\frac{m}{z}$ values as positional inputs.
Recent MS advancements include PowerNovo \cite{petrovskiy2024}, an ensemble of transformer and BERT models for tandem MS peptide sequencing, and a semi-autoregressive transformer framework for rapid sequencing \cite{Zhao2025}. These methods accelerate proteomics but often rely on preprocessed data, contrasting our focus on raw, noisy single-shot spectra from portable aerosol systems for real-time pathogen detection in environmental health monitoring.

\subsection{Proposed MS-DGFormer Architecture}
\begin{figure}[tbp]
\centering
\includegraphics[width=0.95\columnwidth]{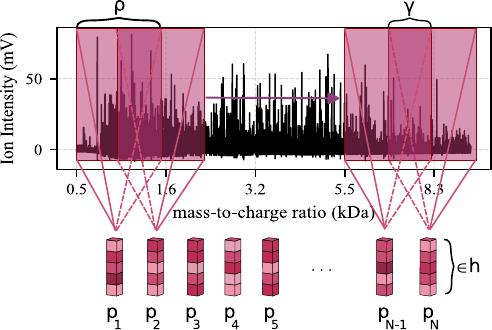}
\caption{The input spectral embedding layer creates a sequence of small overlapping patches from the mass spectrum \(\mathbf{s}\) through one-dimensional convolution filters, transforming the 1D spectrum to 2D sequence.}
\label{fig:fig3}
\end{figure}

\begin{figure*}[tb]
\centering
\includegraphics[width=0.95\textwidth]{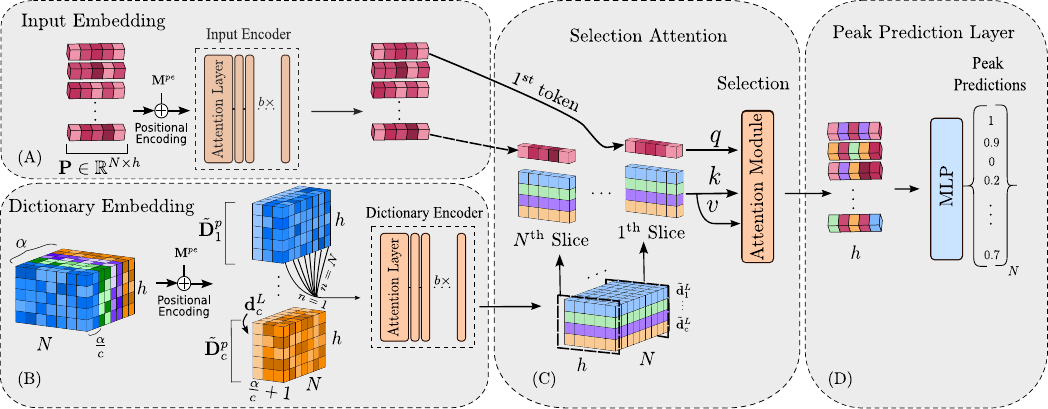} 
\caption{The Mass Spectral Dictionary-Guided Transformer (MS-DGFormer) architecture. (A) Input embedding module. (B) Dictionary Embedding Module. (C) Selection Attention Mechanism. (D) Final peak prediction layer.}
\label{fig:fig4}
\end{figure*}

The Mass Spectral Dictionary-Guided Transformer (MS-DGFormer) processes raw input spectra and denoised dictionary spectra through separate embedding and encoding pathways, enabling robust multi-label classification of biomolecular patterns in noisy aerosol data for real-time public health monitoring (see Fig.~\ref{fig:fig4} for model overview).

\begin{figure*}[tb]
\centering
\includegraphics[width=0.95\textwidth]{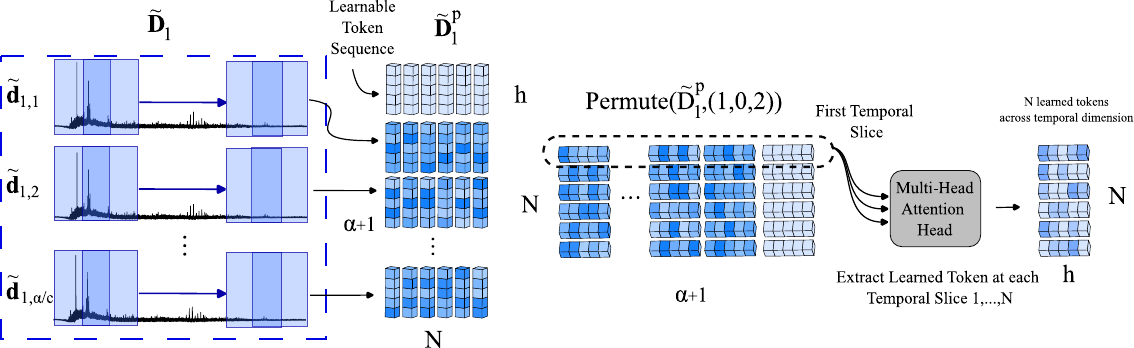}
\caption{The processing of a sub-dictionary is illustrated by exemplifying the first low-rank approximated sub-dictionary \(\tilde{\mathbf{D}}_1 \in \mathbb{R}^{\frac{\alpha}{c} \times l}\). The spectra \([\tilde{\mathbf{d}}_{1,1}, \ldots, \tilde{\mathbf{d}}_{1,\frac{\alpha}{c}}]^{T}\) are transformed into token sequences via convolution with overlapping kernels and \(h\) output channels, yielding \(\tilde{\mathbf{D}}^{p}_{1} \in \mathbb{R}^{\frac{\alpha}{c} \times N \times h}\), where each kernel encodes temporal peak information. A learnable token sequence \(\tilde{\mathbf{d}}^{L}_{1} \in \mathbb{R}^{1 \times N \times h}\) is concatenated with \(\tilde{\mathbf{D}}^{p}_{1}\), forming \(\tilde{\mathbf{D}}^{p}_{1} \in \mathbb{R}^{\left(\frac{\alpha}{c} + 1\right) \times N \times h}\). This tensor is permuted to \(\mathbb{R}^{N \times \left(\frac{\alpha}{c} + 1\right) \times h}\), for attention to be computed independently across the \(N\) temporal positions. The attention mechanism aggregates information across the \(\frac{\alpha}{c} + 1\) sequences at each temporal location. Finally, the learnable tokens \(\tilde{\mathbf{d}}^{L}_{1}\), now enriched with contextual information, are extracted to represent the aggregated temporal information.
}
\label{fig:fig5}
\end{figure*}

\paragraph{Input Embedding}
To handle the intensity vector $\mathbf{s} \in \mathbb{R}^{l}$ and corresponding $\frac{m}{z}$ values $\mathbf{m} \in \mathbb{R}^{l}$, we adapt the "patchification" from Vision Transformers \cite{dosovitskiy2021an} for 1D spectra. Overlapping patches are extracted via 1D convolution with kernel size $\rho$, stride $\gamma$, and $h$ output channels (hidden dimension), yielding $N = \frac{l - \rho}{\gamma} + 1$ embeddings:
$$p_{i,j} = \sum_{k=0}^{\rho-1} w_{j,k} s_{\gamma i + k} + b_j,$$
forming matrix $\mathbf{P} \in \mathbb{R}^{N \times h}$. This convolutional method captures local peaks amid noise, outperforming linear projections by reducing edge artifacts and enhancing robustness (Fig.~\ref{fig:fig3}).
Transformer attention is permutation-invariant, focusing on pairwise token relationships without inherent order. Positional embeddings address this by encoding sequence positions, using methods like fixed sinusoids \cite{vaswani2017attention}, rotary embeddings (RoPE) \cite{Jianlin2023}, or learnable parameters. In mass spectrometry, the $\frac{m}{z}$ vector $\mathbf{m}$ provides intrinsic positional data from time-of-flight. For overlapping patches projected to dimension $h$, we patch $\mathbf{m}$ similarly: $\mathbf{M}_i = \mathbf{m}_{[\gamma (i-1) + 1 : \gamma (i-1) + \rho]} \in \mathbb{R}^{\rho}$, forming $\mathbf{M} \in \mathbb{R}^{N \times \rho}$. A linear projection maps these to $h$:
$$\mathbf{M}^{pe} = \mathbf{M} \mathbf{W}^T + \mathbf{b} \in \mathbb{R}^{N \times h},$$
with $\mathbf{W} \in \mathbb{R}^{h \times \rho}$ and $\mathbf{b} \in \mathbb{R}^{h}$. Adding $\mathbf{M}^{pe}$ to spectral embeddings $\mathbf{P}$ integrates intensity and $\frac{m}{z}$ positions, allowing learned extrapolation to $h$ for better biomolecular pattern recognition in noisy spectra.
\paragraph{Dictionary Embeddings}
Similarly, each spectrum in the denoised dictionary $\tilde{\mathbf{D}} \in \mathbb{R}^{\alpha \times l}$ (Section~\ref{subsec:dictionary_construction}) undergoes convolutional embedding, producing patch sequences $\tilde{\mathbf{d}}^p_i \in \mathbb{R}^{N \times h}$ for $i = 1, \dots, \alpha$, stacked into $\tilde{\mathbf{D}}^p \in \mathbb{R}^{\alpha \times N \times h}$. The same positional embeddings $\mathbf{M}^{pe}$ are added to consistently encode $\frac{m}{z}$ positions. Separate learnable weights for input and dictionary spectra distinguish noisy from denoised features, allowing targeted extraction of clean biomolecular signatures for improved detection accuracy. 
\paragraph{Encoder Blocks}
Both input and dictionary pathways employ transformer encoder blocks based on Vaswani et al. \cite{vaswani2017attention}, featuring multi-head self-attention, an MLP, layer normalization, and residuals for gradient stability. For the input, embeddings $\mathbf{P}$ (with added positional embeddings) are projected to queries ($\mathbf{Q}$), keys ($\mathbf{K}$), and values ($\mathbf{V}$):
$$\mathbf{Q} = \mathbf{P} \mathbf{W}^Q, \quad \mathbf{K} = \mathbf{P} \mathbf{W}^K, \quad \mathbf{V} = \mathbf{P} \mathbf{W}^V,$$
with scaled dot-product attention per head and concatenation via $\mathbf{W}^O$. This captures long-range spectral dependencies essential for noisy data.
For the dictionary, embeddings $\tilde{\mathbf{D}}^p$ (with added positional embeddings) are processed per sub-dictionary $\tilde{\mathbf{D}}^p_i \in \mathbb{R}^{\frac{\alpha}{c} \times N \times h}$. For each sub-dictionary, a learnable sequence $\tilde{\mathbf{d}}^L_i \in \mathbb{R}^{1 \times N \times h}$ is concatenated, the resulting tensor is permuted to $\mathbb{R}^{N \times (\frac{\alpha}{c}+1) \times h}$, and attention is applied slice-wise across sequences at each patch position (Fig.~\ref{fig:fig5}). The \(c\) sequence tokens gather information globally throughout the sub-dictionary, providing aggregated side information for the input encoder's output. Keeping the input and dictionary encoders separate ensures denoised priors guide classification without contaminating raw signals, facilitating precise pathogen identification for biodefense applications.

\begin{figure}[tb]
\centering
\includegraphics[width=0.95\columnwidth]{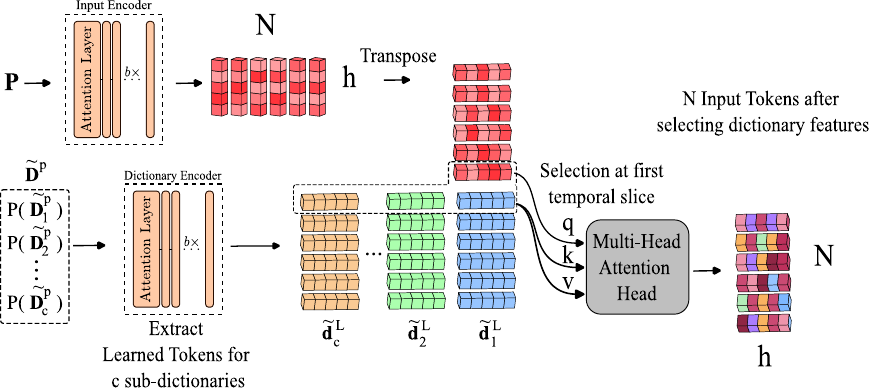}
\caption{An input spectral sequence \(\mathbf{P}\in\mathbb{R}^{N \times h}\) is first encoded by the input encoder. Each sub-dictionary's sequences are permuted to \(\mathbb{R}^{N \times (\frac{\alpha}{c}+1) \times h}\), processed by the dictionary encoder, and the respective learnable token sequences (\(\tilde{\mathbf{d}}^{L}_{i}\)) are extracted. Multi-head attention selects dictionary features for each temporal position.}
\label{fig:fig6}
\end{figure}

\paragraph{Selection Attention}
Following encoding, the input sequence (shape $N \times 1 \times h$) selectively extracts features from the $c$ aggregated sub-dictionary sequences (shape $N \times c \times h$) via a multi-head cross-attention layer (Fig.~\ref{fig:fig6}). The input acts as queries, with sub-dictionaries as keys and values, enabling class-specific feature integration (e.g., prioritizing the relevant low-rank subspace for a given pathogen). A residual connection preserves original input information, enhancing model stability and accuracy in noisy environmental samples.
\paragraph{Peak Prediction}
Known peak locations in training classes form ground-truth $\mathbf{y}_i \in \mathbb{R}^{N}$, where $y_{i,j} = 1$ if class $i$ has a peak at patch $j$. An MLP processes the model output $\mathbf{P}^{out} \in \mathbb{R}^{N \times h}$ for binary predictions:
$$\hat{\mathbf{y}} = \text{sigmoid}\left( \text{ReLU}(\mathbf{P}^{out} \mathbf{W}^{(1)} + \mathbf{b}^{(1)}) \mathbf{W}^{(2)} + \mathbf{b}^{(2)} \right),$$
with $\mathbf{W}^{(1)} \in \mathbb{R}^{h \times \phi}$, $\mathbf{b}^{(1)} \in \mathbb{R}^{\phi}$, $\mathbf{W}^{(2)} \in \mathbb{R}^{\phi \times 1}$, and $\mathbf{b}^{(2)} \in \mathbb{R}$. This outputs probabilities per patch, supporting multi-label classification for rapid biomolecular threat detection in aerosols. The final predicted class $\hat{c}$ is the one with maximum cosine similarity to ground truth vectors $\mathbf{y}_c$ for $c \in \{1, \dots, 5\}$:
$$\hat{c} = \arg\max_{c} \frac{\hat{\mathbf{y}} \cdot \mathbf{y}_c}{\|\hat{\mathbf{y}}\| \|\mathbf{y}_c\|}.$$
Training optimizes binary cross-entropy between $\hat{\mathbf{y}}$ and $\mathbf{y}$, while class predictions inform evaluation metrics: accuracy, precision, recall, and F1 score.

\section{Results}
\subsection{Competing Models}
To evaluate our model's performance, we benchmarked it against recurrent baselines and a dictionary-ablated variant, focusing on sequence modeling for noisy mass spectra in pathogen detection. To ensure a fair comparison, we strive to maintain consistency in model parameters where possible; however, due to architectural differences, exact parameter matching is not always feasible.

Recurrent Neural Network (RNN)\cite{Elman1990}, Long Short-Term Memory (LSTM)\cite{Hochreiter1997}, and Bidirectional LSTM (biLSTM)\cite{Schuster_bilstm} models retained our input embedding and peak prediction layers for consistent processing and output. Positional embeddings were omitted, as RNNs and LSTMs inherently capture sequential order. The dictionary, input encoder, dictionary encoder, and selection attention were replaced with RNN, LSTM, or biLSTM blocks. This evaluation directly compares recurrent vs. transformer-based sequence handling in biomolecular classification.

To assess the dictionary's impact in our model, we trained a model without the dictionary embedding, encoder, and selection attention (MS-Former), which halved the total parameter count. Essentially this model is a standard transformer. For fair comparison, we trained variants with 3 input encoder layers (MS-Former-3; 4.13M parameters) to match our core architecture and 7 layers (MS-Former-7) to approximate total parameters (8-9M across models), isolating the dictionary's role in enhancing accuracy for environmental health applications.

\subsection{Model and Dictionary Hyperparameters}
Experiments used convolutional embedding window size $\rho = 100$ and overlap $\gamma = 50$ (50\%), yielding $N = 1765$ patches for spectra of length $l = 88300$. Corresponding $\frac{m}{z}$ values were patched identically. Patches mapped to hidden dimension $h = 256$.
Multi-head attention ($n_{\text{heads}} = 8$, $d_k = 32$) was applied in the input encoder's self-attention, dictionary encoder's slice attention, and selection attention. Attention MLP intermediate dimension was 2048; peak prediction MLP was $\phi = 512$. Both encoders had $L = 3$ layers.
Dictionary $\mathbf{D}$ comprised $\alpha = 32$ sequences from 4 positive classes (\textit{B. globigii}, \textit{E. coli}, insulin, ubiquitin; 8 per class), excluding dust. Limited by single 4070 GPU memory, it was denoised via sub-dictionary SVD with rank $r = 2$, providing efficient side information for robust pathogen detection in portable systems. 

Our model and competing models were trained for \(300\) epochs with batch size of \(8\), a learning rate of \(10^{-4}\) with \(10\%\) warm-up, and a cosine annealing decay. There were two types of regularization implemented: neuron dropouts placed similarly to \cite{vaswani2017attention} with \(0.1\) dropout probability, and binary cross-entropy label smoothing. This type of label smoothing treats each \(1\) as \(0.9\) and each \(0\) as \(0.1\), so the model does not become overconfident in its peak predictions. 

\subsection{MS-DGFormer Evaluation}
\begin{figure}[tb]
\centering
\includegraphics[width=0.95\columnwidth]{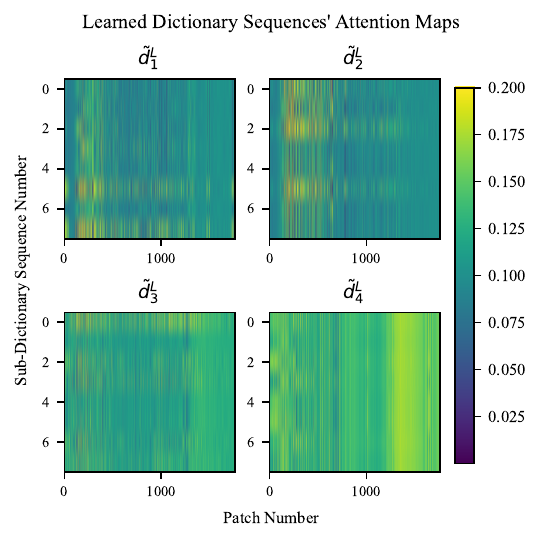}
\caption{Each class's sub-dictionary attention maps averaged across heads are shown. The larger attention scores are located at each class's true peak locations showing the dictionary's efficacy for feature extraction.}
\label{fig:fig7}
\end{figure}

We begin by examining the results obtained from MS-DGFormer before proceeding to a comparison with competing models. To understand the features captured in each sub-dictionary, we visualize the attention maps derived from each learnable sequence, \(\tilde{\mathbf{d}}^{L}_{i}\). Figure~\ref{fig:fig7} displays the average attention scores, revealing that patches containing spectral peaks receive higher attention scores compared to those without. Thus, \(\tilde{\mathbf{d}}^{L}_{i}\) effectively focuses on the peaks found within the \(i^{\rm th}\) sub-dictionary.

\begin{figure}[tb]
\centering
\includegraphics[width=0.95\columnwidth]{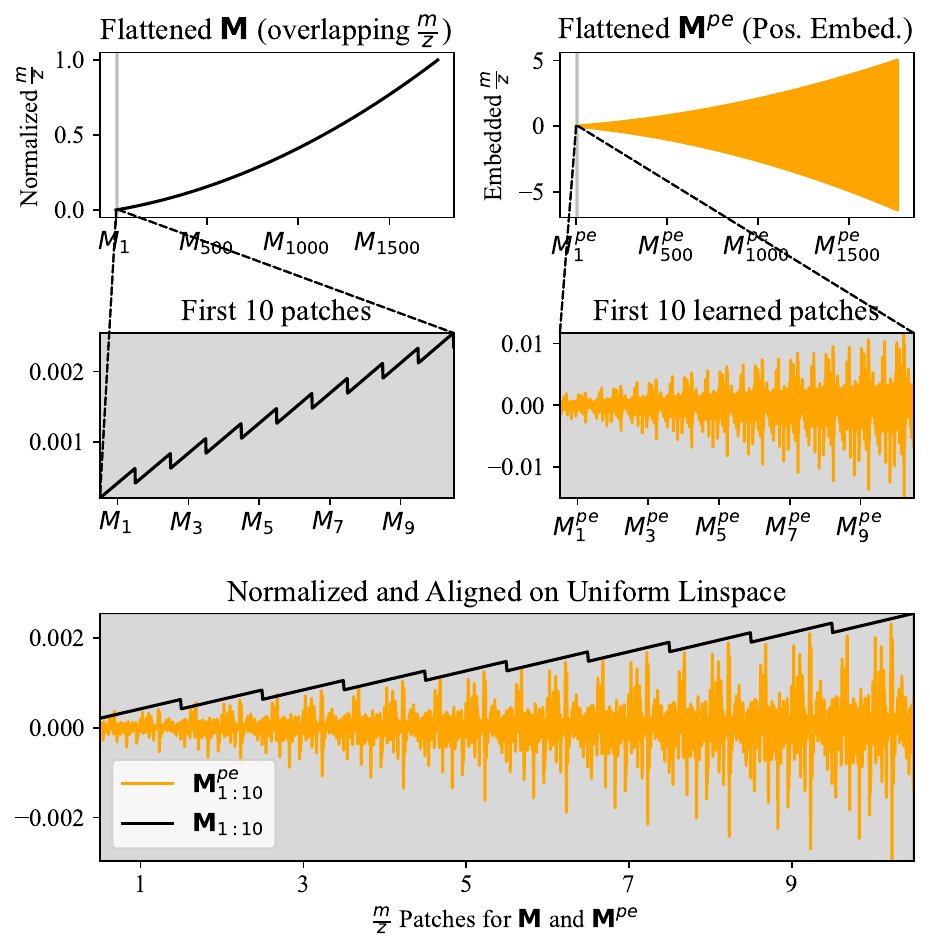}
\caption{Top: The overlapping patches of \(\frac{m}{z}\) values are flattened to \(\mathbb{R}^{N\rho}\) (left). \(\mathbf{M}\) is embedded via a learnable linear layer producing \(\mathbf{M}^{pe}\) (right). Middle: The first 10 patches of \(\mathbf{M}\) and \(\mathbf{M}^{pe}\). Bottom: The patches are normalized and plotted on the same linspace.}
\label{fig:fig8}
\end{figure}

\begin{figure}[tb]
\centering
\includegraphics[width=0.95\columnwidth]{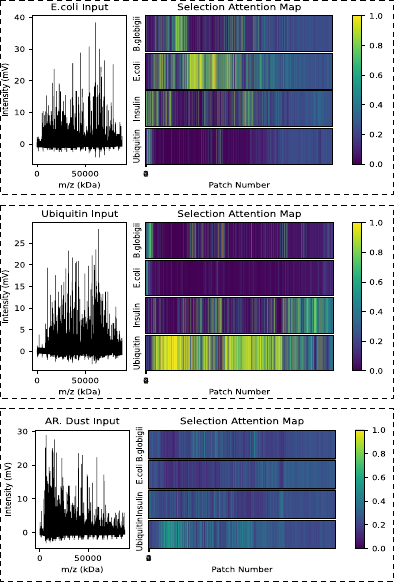}
\caption{Top: \textit{E.coli} input spectrum and the attention map from the selection attention mechanism. The attention scores are larger for \textit{E.coli}, capturing the learning of the selection mechanism. Middle: Ubiquitin input spectrum and its corresponding selection attention map. Bottom: Arizona Road Dust and its corresponding selection attention map showing no dominant features selected from a single class.}
\label{fig:fig9}
\end{figure}

Next, to gain insight into how the model processes the raw overlapping \(\frac{m}{z}\) values and maps them to its embedding space, we extract and visualize the positional embeddings. For visualization purposes, we flatten both the raw \(\frac{m}{z}\) values and the corresponding positional embeddings. Specifically, the raw \(\frac{m}{z}\) matrix \(\mathbf{M} \in \mathbb{R}^{1765 \times 100}\) is flattened to a vector in \(\mathbb{R}^{176500}\), and the positional embeddings \(\mathbf{P} \in \mathbb{R}^{1765 \times 256}\) are flattened to a vector in \(\mathbb{R}^{451840}\). We then plot these flattened vectors, focusing on the segments corresponding to the first 10 patches, as illustrated in Fig.~\ref{fig:fig8}. From this figure, it is evident that the model preserves the structural characteristics of the raw \(\frac{m}{z}\) values, such as their overlapping nature (with 50\% overlap between adjacent patches), while mapping them to a higher-dimensional embedding space (from 100 to 256 dimensions per patch). Additionally, the amplitude of the positional embeddings is adjusted to a range of approximately \(\pm5\), aligning with the intensity scales observed in the mass spectra. To demonstrate the alignment between the positional embeddings and the original \(\frac{m}{z}\) values, we normalize both to a common amplitude range and plot them on a uniform linear space. Fig.~\ref{fig:fig9} illustrates this alignment for each patch, highlighting the model's ability to encode positional information effectively.

Finally, we visualize the attention maps within the selection attention head to understand how the model selects features from the sub-dictionaries based on the input spectrum class. To achieve this, we input a representative bacteria, protein, and noise spectrum into the model and extract the attention map from the selection attention head for visualization. Fig.~\ref{fig:fig9} shows the attention scores across the sub-dictionaries for each input spectrum. From this figure, it is evident that for each input spectrum, the attention scores are significantly higher for the sub-dictionary corresponding to the same class. However, for the Arizona Road Dust class, the attention scores are more dispersed and lower in magnitude. This behavior is expected because the Arizona Road Dust spectra primarily contain noise peaks, which do not align well with the denoised spectra represented in the sub-dictionaries.

\subsection{Competing Model Comparisons}

Having demonstrated the spectral features captured by MS-DGFormer, we now proceed to evaluate its performance against competing models using standard classification metrics. Each model is assessed on the test set, with performance measured using micro and macro accuracy, precision, recall, and F1-score. Table III presents the macro metrics, where MS-DGFormer achieves the highest scores across all metrics, despite having fewer parameters than most other models, with the exception of MS-Former-\(3\). Notably, the biLSTM-based model outperforms MS-Former-\(7\) but still falls short of MS-DGFormer’s performance.

For brevity, we focus on the micro F1-score in our analysis, as it effectively balances the trade-off between false positives and false negatives, which is critical for classification tasks. Table IV displays the micro F1-scores for each class, with MS-DGFormer consistently achieving the highest F1-score across all classes. In contrast, the other models exhibit significant performance degradation on the Arizona Road Dust class. This is particularly concerning because dust-like particles are commonly encountered in environmental conditions, and misclassifying them as biological agents could result in a high rate of false alarms.
\begin{table}[h]
\centering
\caption{Macro Metrics Across All Classes.}
\renewcommand{\arraystretch}{1.2} 
\resizebox{\columnwidth}{!}{
\begin{tabular}{|c|c|c|c|c|c|}
\hline
\multirow{2}{*}{Model} & \multicolumn{5}{c|}{Macro Metrics} \\ \cline{2-6}
    & Accuracy & Precision & Recall & F1 & Params  \\ \hline
RNN-\(6\)         & 0.560  & 0.832 & 0.560 & 0.491 & 9.52M\\ \hline
LSTM-\(4\)        & 0.679  & 0.821 & 0.679 & 0.641 & 9.50M\\ \hline
BiLSTM-\(6\)      & 0.939  & 0.916 & 0.939 & 0.915 &9.48M \\ \hline
MS-Former-\(3\)   & 0.709  & 0.845 & 0.709 & 0.664 & 4.13M\\ \hline
MS-Former-\(7\)   & 0.862  & 0.876 & 0.862 & 0.824 & 9.39M\\ \hline
MS-DGFormer       & \textbf{0.983} & \textbf{0.982}& \textbf{0.983}    & \textbf{0.982}  & 8.36M\\ \hline
\end{tabular}
}
\end{table}    

\begin{table}[h]
\centering
\caption{Micro F1 Scores For Each Class}
\label{tab:f1_scores}
\renewcommand{\arraystretch}{1.2} 
\resizebox{\columnwidth}{!}{
\begin{tabular}{|c|c|c|c|c|c|}
\hline
\rule{0pt}{2.5ex} \textbf{Model} & \textbf{A.R.Dust} & \textbf{B.globigii} & \textbf{E.coli} & \textbf{Insulin} & \textbf{Ubiq.} \\ \hline
\rule{0pt}{2.5ex} RNN-6   & 0.278  & 0.045  & 0.408  & 0.795  & 0.926 \\ \hline
\rule{0pt}{2.5ex} LSTM-4  & 0.331  & 0.623  & 0.374  & 0.906  & 0.972 \\ \hline
\rule{0pt}{2.5ex} BiLSTM-6 & 0.736  & 0.926  & 0.954  & 0.976  & 0.983  \\ \hline
\rule{0pt}{2.5ex} MS-Former-3  & 0.369  & 0.328  & 0.880  & 0.787  & 0.952  \\ \hline
\rule{0pt}{2.5ex} MS-Former-7  & 0.553  & 0.666  & \textbf{0.984}  & 0.921  & \textbf{0.994}  \\ \hline
\rule{0pt}{2.5ex} MS-DGFormer & \textbf{0.949} & \textbf{0.991} & 0.979 & \textbf{0.987} & \textbf{0.994} \\ \hline
\end{tabular}
}
\end{table}

\begin{table*}[t]
\caption{Inference Performance Metrics for Different Models and Batch Sizes}
\label{tab:performance_metrics}
\renewcommand{\arraystretch}{1.2} 
\resizebox{\textwidth}{!}{%
\begin{tabular}{c|ccc|ccc|ccc} 
\toprule
Model & \multicolumn{3}{c|}{Batch Size 1} & \multicolumn{3}{c|}{Batch Size 4} & \multicolumn{3}{c}{Batch Size 8} \\
\cline{2-10}
\renewcommand{\arraystretch}{1.2} 

& Mean (ms) \(\downarrow\) & Std (ms) \(\downarrow\) & Spectra/s \(\uparrow\) & Mean (ms) \(\downarrow\) & Std (ms) \(\downarrow\) & Spectra/s \(\uparrow\) & Mean (ms) \(\downarrow\) & Std (ms) \(\downarrow\) & Spectra/s \(\uparrow\)\\
\midrule

RNN-\(6\)  & 125.32 & 3.36  & 7.98 & 156.30  &  4.03  & 25.59  & 221.37 & 5.94  & 36.14 \\

LSTM-\(4\) & 108.62 & 3.31  & 9.21  & 116.619  & 4.79  & 34.30  & 194.51 & 6.79  & 41.13  \\

BiLSTM-\(6\)  & 156.38 & 2.56 & 6.39 & 263.82 & \textbf{1.46} & 15.16 & 223.67 & 4.18 & 35.77 \\

MS-Former-\(3\) & \textbf{11.88} & 2.55 & \textbf{84.15} & \textbf{34.70} & 3.31 & \textbf{115.27} & \textbf{62.87} & 4.91 & \textbf{127.25} \\

MS-Former-\(7\) & 23.46 & 2.32 & 42.63 & 74.31 & 4.18 & 53.83 & 141.19 & 3.08 & 56.66 \\

MS-DGFormer & 72.27 & 3.76 & 13.84 & 95.66 & 2.85 & 41.81 & 127.17 & \textbf{2.58} & 62.91 \\

MS-DGFormer-E & \textbf{12.31} & \textbf{1.67} & \textbf{81.23} & \textbf{35.98} & 3.70 & \textbf{111.16} & \textbf{66.33} & 3.92 & \textbf{120.60} \\
\bottomrule
\end{tabular}%
}
\end{table*}

\subsection{Computational Efficiency}
Real-time field analysis demands rapid processing of continuous spectral streams, where both parameter count and hardware requirements are critical. Our design improves efficiency by separating the dictionary embedding/encoder from the input embedding/encoder. The dictionary sequences remain constant during training and are encoded independently of the input spectrum. Only the learned sequence per sub-dictionary is used: \(\alpha\) sequences enter the dictionary encoder during training, but only \(c\) sequences are passed to the selection attention head.

Because the dictionary input is spectrum-independent, we precompute and store the \(c\) learned sub-dictionary sequences, removing the dictionary embedding layer, dictionary encoder, and dictionary itself from the inference model. This yields an efficient variant, MS-DGFormer-E, where only the pre-trained weights of the remaining components are loaded. The \(c\) sequences are stored in memory and fed directly into the selection attention head, cutting parameters from \(8.36 \times 10^6\) to \(4.39 \times 10^6\) without performance loss.

This architecture also scales efficiently. Adding more sequences to sub-dictionaries during training does not affect inference time, as the number of sequences used at inference, \(c\), remains fixed. If \(\theta\) new classes are added, the training dictionary expands to \(\frac{\alpha}{c} \theta + \alpha\) sequences, yet inference still requires only \(c + \theta\) sequences, with \(c \ll \frac{\alpha}{c}\).

We evaluated inference speed and spectral throughput on batches of size 1, 4, and 8, averaging 100 runs after 10 warm-up runs. As shown in Table~V, MS-DGFormer-E achieves nearly a \(2\times\) increase in mean inference speed and more than a \(2\times\) increase in throughput over the full MS-DGFormer. The only model with comparable inference time is MS-Former-3, due to its similar parameter count, but its classification performance is significantly lower. While results are hardware-specific, the relative efficiency gains should generalize across platforms.

\section{Discussion}
We have introduced an approach capable of classifying individual noisy mass spectra without the need to collect several spectra of the same class to create an averaged spectrum. The mass spectra are turned into sequences of small overlapping patches enabling attention mechanisms to capture peak locations. Although, this alone is not enough to accurately classify the spectral input due to intense noise. We notice that if we do have a batch of spectra all from the same class, the SVD is able to significantly reduce noise and reveal true spectral peaks. However, it is unlikely this will naturally occur in the environment. We construct a dictionary composed of training samples from each class, clustered together to resemble sub-dictionaries, and perform the SVD on the sub-dictionaries to be used as side-information. One learnable (randomly initialized) sequence is concatenated to each sub-dictionary to capture features for their respective sub-dictionary for both effectiveness and efficiency. Further, our model processes the input spectrum and dictionary through separate encoders allowing the features to be learned independently, with another attention head capable of selecting specific information from the dictionary. In this manner, only the learned dictionary sequences are required for inference, and the dictionary components can be removed from the model. This significantly improves inference speed and spectral throughput since nearly half of the model parameters reside in the dictionary components. Finally, our MS-DGFormer is compared against competing sequential models and achieves the highest micro and macro classification metrics with drastically faster inference metrics.

Overall, our proposed architecture, MS-DGFormer, addresses several challenges that arise when transitioning MALDI-MS systems from laboratory to environmental settings. Traditional MALDI-MS workflows require extensive sample preparation and pre-processing; our approach eliminates much of this burden through autonomous aerosol sampling, shifting the workload to computational post-processing and machine learning inference. Rather than relying on multi-shot spectral averaging to obtain clean spectra, we leverage SVD-denoised sub-dictionaries to provide rich side information for accurate classification of raw, minimally processed spectra.

False positive classification is a particular challenge in environmental sampling, where negative classes such as dust are abundant. MS-DGFormer demonstrates robust performance in these scenarios, effectively mitigating misclassification.

Our results highlight the potential of real-time pathogen detection using portable aerosol MALDI-MS systems, paving the way for field-deployable solutions that can transform environmental monitoring, biological threat detection, and efforts to reduce disease spread.

\section*{Declaration of competing interest}
W.B., and M.M. have competing interests. W.B. is the President and CEO of Zeteo Tech, Inc. M.M. is the Vice President of Research and CTO at Zeteo Tech, Inc.

\section*{Acknowledgments}
This work was supported in part by the National Institute of Health under award number T32GM142603 and in part by Zeteo Tech Inc. Due to privacy or ethical restrictions, the data from the study is only available upon request from the corresponding author. Code is available upon request from the corresponding author.

\section*{Author Contributions}
Conceptualization, K.R., G.A., M.M. and W.B; methodology,  K.R., G.A., M.M. and W.B; software, K.R; validation, G.A., M.M., W.B.; formal analysis, K.R., G.A.; investigation, K.R., G.A., M.M. and W.B; resources, K.R., G.A., M.M. and W.B; data curation, K.R, M.M, W.B; writing---original draft preparation, K.R., G.A; writing---review and editing, G.A.,  M.M. and W.B.; visualization, K.R.; supervision, G.A., M.M. and W.B.; project administration, G.A., M.M. and W.B.; funding acquisition, G.A., M.M. and W.B. All authors have read and agreed to the published version of the manuscript.

\section*{Declaration of generative AI and AI-assisted technologies in the manuscript preparation process}
During the preparation of this work the author(s) used Grok xAI in order for grammar correction and readability. After using this tool/service, the author(s) reviewed and edited the content as needed and take(s) full responsibility for the content of the published article.



\bibliographystyle{elsarticle-num} 
\bibliography{references}






\end{document}